\documentclass[sigconf]{acmart}
\AtBeginDocument{%
  }

\copyrightyear{2025}
\acmYear{2025}
\setcopyright{acmlicensed}\acmConference[CIKM '25]{Proceedings of the 34th ACM International Conference on Information and Knowledge Management}{November 10--14, 2025}{Seoul, Republic of Korea}
\acmBooktitle{Proceedings of the 34th ACM International Conference on Information and Knowledge Management (CIKM '25), November 10--14, 2025, Seoul, Republic of Korea}
\acmDOI{10.1145/3746252.3761547}
\acmISBN{979-8-4007-2040-6/2025/11}

\usepackage{times}
\usepackage{latexsym}

\usepackage{multirow}
\usepackage{makecell}
\usepackage{balance}

\usepackage[T1]{fontenc}
\usepackage{graphicx}
\usepackage{xcolor}
\usepackage{ulem}
\usepackage{amsmath}

\newtheorem{definition}{\sc Definition}

\newcommand{\stitle}[1]{\vspace*{0.5em}\noindent{\bf #1.\/\;}}
\newcommand{\modelname}{\textsf{MHSNet}}
\newcommand{\usedllm}{\textsf{BGE-M3}}

\begin{document}

\title[LLM-Enhanced Hierarchical MoE Representation Network for Duplicate Resume Detection]{MHSNet: An MoE-based Hierarchical Semantic Representation Network for Accurate Duplicate Resume Detection with Large Language Model}


\author{Yu Li}
\authornote{Both authors contributed equally to this research.}
\affiliation{
  \institution{
  Hangzhou Dianzi University
  }
  \city{Hangzhou}
  \country{China}}
\email{liyucomp@hdu.edu.cn}

\author{Zulong Chen}
\authornotemark[1]
\affiliation{
  \institution{Alibaba Group}
  \city{Hangzhou}
  \country{China}
}
\email{zulong.czl@alibaba-inc.com}

\author{Wenjian Xu}
\affiliation{
  \institution{
    Zhejiang University of Science and Technology
  }
  \city{Hangzhou}
  \country{China}}
\email{wenjian.xwj@zust.edu.cn}

\author{Hong Wen}
\authornote{Corresponding Author: Yuyu Yin, Hong Wen.}
\affiliation{
  \institution{
  Affiliation
  }
  \city{Hangzhou}
  \country{China}
}
\email{dreamonewh@gmail.com}

\author{Yipeng Yu}
\affiliation{
 \institution{
 Taotian, Alibaba Group
 }
 \city{Hangzhou}
 \country{China}}
 \email{linxin.yyp@alibaba-inc.com}

\author{Man Lung Yiu}
\affiliation{
  \institution{
  Department of Computing, Hong Kong Polytechnic University
  }
  \city{Hong Kong}
  \country{China}}
\email{csmlyiu@comp.polyu.edu.hk}

\author{Yuyu Yin}
\authornotemark[2]
\affiliation{%
  \institution{Hangzhou Dianzi University}
  \city{Hangzhou}
  \country{China}}
\email{yinyuyu@hdu.edu.cn}


\begin{abstract}
 To maintain the company's talent pool, recruiters need to continuously search for resumes from third-party websites (e.g., LinkedIn, Indeed). However, fetched resumes are often incomplete and inaccurate. To improve the quality of third-party resumes and enrich the company's talent pool, it is essential to conduct duplication detection between the fetched resumes and those already in the company's talent pool.
Such duplication detection is challenging due to the semantic complexity, structural heterogeneity, and information incompleteness of resume texts.
To this end, we propose \modelname{}, an multi-level identity verification framework that fine-tunes \usedllm{} using contrastive learning. With the fine-tuned \usedllm{}, \modelname{} generates multi-level sparse and dense representations for resumes, enabling the computation of corresponding multi-level semantic similarities.
%
%we propose a multi-level identity verification framework \modelname{}, which fine-tunes \usedllm{} with contrastive learning. 
%With fine-tuned \usedllm{}, \modelname{} calculates multi-level sparse and dense representations for resumes, and corresponding multi-level semantic similarity are obtained.
Moreover, the state-aware Mixture-of-Experts (MoE) is employed in \modelname{} to handle diverse incomplete resumes.
Experimental results verify the effectiveness of \modelname{}.
\end{abstract}



%\ccsdesc[500]{Do Not Use This Code~Generate the Correct Terms for Your Paper}
%\ccsdesc[300]{Do Not Use This Code~Generate the Correct Terms for Your Paper}
%\ccsdesc{Do Not Use This Code~Generate the Correct Terms for Your Paper}
%\ccsdesc[100]{Do Not Use This Code~Generate the Correct Terms for Your Paper}



\maketitle

%\begin{CJK}{UTF8}{gbsn}
\section{Introduction} \label{sec:intro}
\begin{figure*}[!hpt]
\centering
\includegraphics[width=0.8\textwidth]{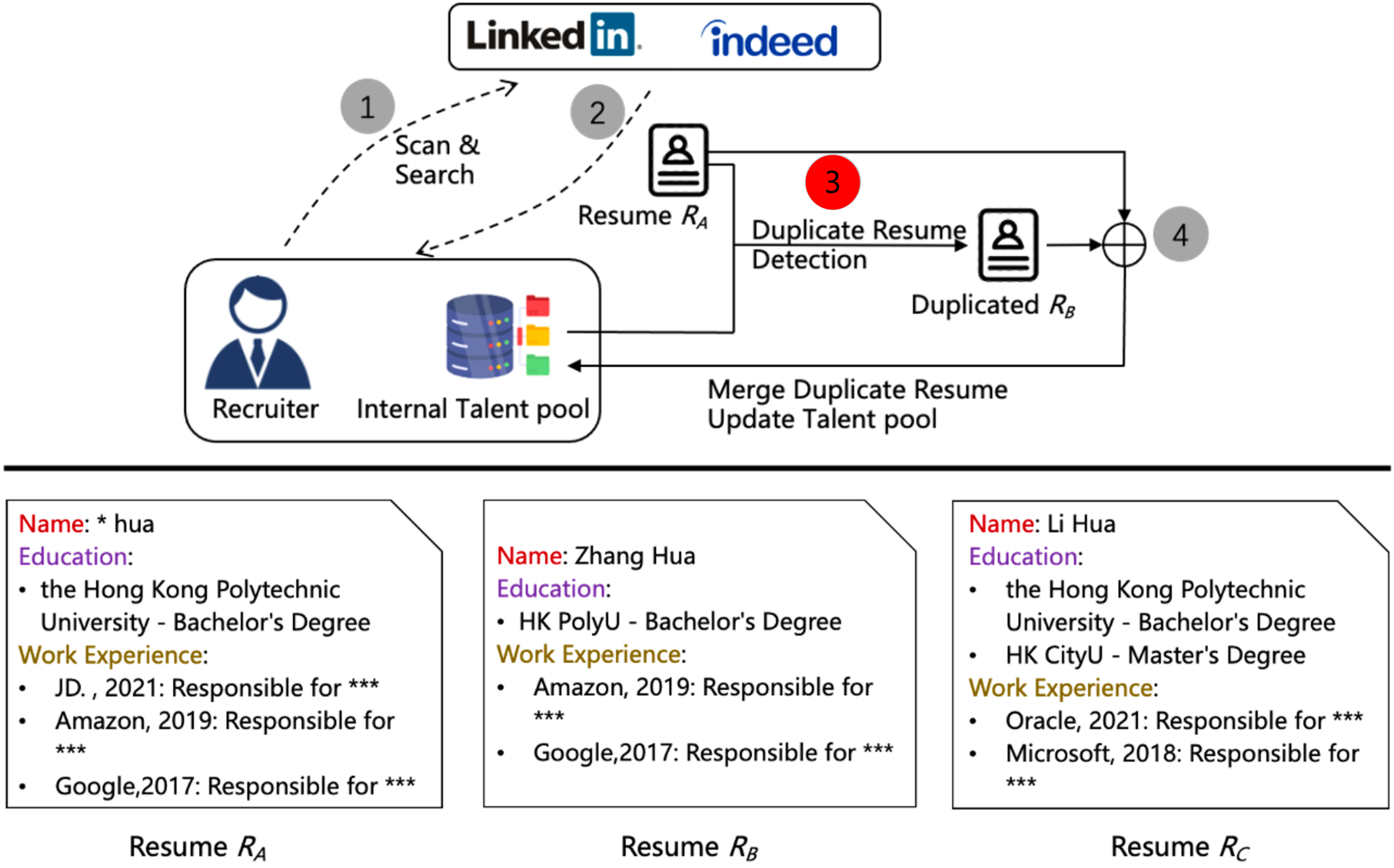}
  \caption{Example case of duplicate resume detection. Resumes $R_A$ and $R_B$ are duplicate while $R_A$ and $R_C$ represent different candidates.}
\vspace{-10pt}
\label{fig:application}
\end{figure*}

Duplicate Resume detection refers to the process of determining whether two given resumes belong to the same individual.
This is challenging because those resumes may have missing information or inconsistent wording in their descriptions.

As shown in Figure~\ref{fig:application}, when recruiters are looking for candidates, they may scan and search from third-party websites (e.g., LinkedIn, Indeed). 
After obtaining the resumes from the third-party websites, recruiters will match them with the company's internal resume pool to identify duplicate resumes and merge the duplicate resumes into a more complete resume for subsequent job matching. 
However, resumes provided by third parties are often incomplete, and the same candidate may also upload multiple different resumes at different times. 
According to our statistics over a random sample of 6,000 resumes, we found that $99\%$ contained inaccurate names, $1.5\%$ lacked  educational information, and
$13\%$ were with incomplete work experience.  
%
%{\color{red} 每年公司有几十万次面试（校招+社招），每次面试需要耗费巨大的人力物力，通过简历查重操作，可以给公司节省百万经费的同时，极大提高招聘效率}
%在信息不完整的简历中进行duplicate detection 对公司招聘有着巨大的作用。因为每年公司有几十万次面试（校招+社招），每次面试需要耗费巨大的人力物力，通过简历查重操作，可以给公司节省百万经费的同时，极大提高招聘效率。
Duplicate detection in resumes with incomplete information plays a pivotal role in corporate recruitment. Given that companies conduct hundreds of thousands of interviews annually (via campus and social recruitment channels), each requiring substantial human and material resources, implementing resume duplication detection can save millions in operational costs while significantly enhancing recruitment efficiency.

%We observe that, in a random sample of 6,000 third-party resumes, $99\%$ had inaccurate names, $1.5\%$ had missing educational information, and $13\%$ had missing work experience. 

Although natural language processing has been employed to improve resume-job matching~\cite{wu2024exploring,qin2018enhancing,yingpeng2024,zhipeng2023}, current work primarily focuses on matching resumes with job requirements, which cannot be used to address the problem of resume de-duplication.%{\color{red} 扩展说一下目前工作的不足}
%(i.e., determining whether two resumes belong to the same individual).

Duplicate resume detection is complex. 
As shown in Figure~\ref{fig:application}, although resumes $R_A$ and $R_C$ have the same undergraduate school in their educational background, their work experience do not match (e.g., JD in 2021 vs. Oracle in 2021). As comparison, although the school names in resumes $R_A$ and $R_B$ appear different, domain knowledge reveals that \textit{HK PolyU} is simply an abbreviation of \textit{the Hong Kong Polytechnic University}. Although the work experiences in $R_A$ and $R_B$ are not the same, it can be seen that $R_B$ is a resume registered a few years ago, and thus the work experience in 2021 is missing. Therefore, $R_A$ and $R_B$ refer to the same candidate, meaning that they are duplicate resumes.

There are many challenges in duplicate resume detection:
%$\mathcal{C}_1$ \textit{Semantic Nature of Resumes} : Resumes are highly semantic-rich documents, containing numerous proper nouns and specialized terms. The semantic representation of resumes is of utmost importance.
$\mathcal{C}_1$, resumes are highly semantic-rich documents, making their semantic representation crucial. Additionally, resume data is imbalance, with negative sample (e.g., $R_A$ and $R_C$ in Figure~\ref{fig:application})  being particularly rare;
%
%$\mathcal{C}_2$ \textit{Importance of Structured Fields}: Resumes are semi-structured texts, comprising numerous key fields such as educational background and work experience. Understanding these fields is crucial. \textit{Global and Local Judgments}: To determine whether two resumes belong to the same individual, it is necessary to assess their similarity from multiple perspectives, including both global and local views. When evaluating, one typically starts with the basic information. If it is highly similar, it is likely that the resumes belong to the same person. If not, further examination of other details is required.
$\mathcal{C}_2$, resumes are typically long, semi-structured texts comprising numerous fields, such as educational and work experience. 
Both local and global similarities play a crucial role in duplicate detection;
%
%$\mathcal{C}_3$ \textit{Diverse Missing Fields and Scarce Samples}: Resumes may have various types of missing fields. Names may be masked, and educational or work experiences may be absent, which can affect the judgment of whether two resumes belong to the same person. Additionally, resume samples are scarce, especially negative samples (i.e., cases where resumes are mistakenly identified as belonging to the same person but actually do not) are extremely rare and difficult to construct.
$\mathcal{C}_3$, different resumes may have varying missing fields, making it challenging for a single network to handle all types of missing information effectively. 

Therefore, to conduct accurate duplicate resume detection for recruiters, this paper proposes an multi-level duplicate detection framework \modelname{}.
In detail, 
to address $\mathcal{C}_1$, \modelname{} employs contrastive learning and element-wise data augmentation techniques to fine-tune embedding model \usedllm{}~\cite{bgem3}.
%while incorporating rich semantic representations, including both dense and sparse representations.
To address $\mathcal{C}_2$, \modelname{} calculates similarities between structured, semi-structured and full resumes to capture both global and local similarities between resumes.
To address $\mathcal{C}_3$, \modelname{} utilizes Mixture-of-Experts (MoE) with the gating network directing the data to different expert modules according to the structured fields state in resumes.
%approach. Depending on whether the content of each field in the resume is missing, the gating network directs the data to different expert modules to address the issue.
%
The main contributions could be summarized as follows:
\begin{itemize}
    \item To our best knowledge, we are the first to conduct duplication detection over incomplete resumes.
    \item We propose \modelname{} that leverages LLM, semantic computation, and MoE to conduct duplicate detection.
    \item We conduct extensive experiments on real-world resume datasets, and the experimental results show the superiority of our model.
\end{itemize}

\iffalse
简历同人判定是指根据信息有缺失、描述用词不对齐的简历来判定给定的两份简历是否表示同一个人。
如图~\ref{fig:application} 所示，公司的招聘人员在寻找候选人时，会从第三方网站（如LinkedIn）搜索简历。
第三方提供的简历往往是不完整的，或者同一候选人也可能会在不同的时间上传多份不同的简历。
招聘人员拿到第三方网站的简历后，会跟公司内部的人才库进行匹配，识别同人简历，并将第三方引入的简历和人才库的简历合并成一份更完整的简历用来进行后续的岗位匹配等。
然而，第三方提供的简历往往是不完整不可靠的.
例如，随机抽取 6000 份三方简历，其中虚假姓名占比 99\%，教育信息为空的占比 1.5\%，工作信息为空的占比 13\%. 
如图~\ref{fig:application} 所示，三份来自第三方的简历都是不完整的，其中resume A和resume C虽然教育经历中有完全相同的本科学校，但是从硕士学历及工作经历分析可知这两份简历不代表同一候选人。resume A和resume B虽然学校名字看着不同，但是根据领域知识，可知PolyU是缩写，虽然工作经历不完全相同，但是可以看出resume B是前几年登记的经历，所以缺少2021年那段工作经历。因此，resume a和resume b是表示同一个候选人，即为duplicate resume。

虽然自然语言处理已经被用于更好的匹配简历与工作~\cite{}，但是目前的工作主要关注计算准确的简历与工作需求之间的关联性，无法用来解决同人判定的问题。
简历同人判定有许多挑战：
$\mathcal{C}_1$简历语义性,简历是一种语义特别丰富的材料，里面设计到很多专有名词以及特殊词语，简历的语义表征极为关键
$\mathcal{C}_2$ 结构化字段重要性,简历是一种半结构化的文本，里面有大量的丰富字段，比如：教育经历、工作经历等。这些经历的理解是非常重要的;全局\&局部判断,要判断两份简历是不是一个人，需要从全局和局部等多个视角来看两个简历的相似度。人在评估的时候也是要先看一看基础信息，如果基础信息非常相似大概率是一个人。如果不是，在看看其他资料.
$\mathcal{C}_3$ 简历部分字段缺失的类型繁多，可能姓名进行了打码，可能教育经历/工作经历缺失，一定程度上影响简历同人的判断，
同时，简历样本稀缺,负样本（即判定为同人而真实却不是同人）是极其稀少，而且难以构造的

因此，为解决上述挑战，本文提出同人判定框架\modelname{}。

For $\mathcal{C}_1$, 采用对比学习、元素级的数据增强的技术来预训练 bge-m3,并同时考虑引入丰富的语义表征，包括 dense 表征和 sparse 表征.
For $\mathcal{C}_2$,分别建模全局简历相似度和局部简历相似度,利用 LLM 来抽取结构化的简历文本信息，例如：基础信息、教育信息、工作信息,建模结构化简历内容的相似度.
For $\mathcal{C}_3$, 采用 MoE 来根据简历的各个字段内容是否缺失，来指导门控网络路由到不同的专家来解决.
\fi
%\end{CJK}
%\input{3-Problem}
\section{Method}\label{sec:method}

\iffalse
4.1: system overview
这个模型是做什么的。

2：输入是什么：chunk符号，strcu符号，resume 
1：offline：模型微调。motivation+总分技术+特别重要的技术展开的细节。微调bge-m3，为了更好地进行饿不饿钉钉。但是样本不均衡，负样本很少---，通过mask机制，结合对比学习，微调bge-m3.
从简历中抽取embedding，
3：multi level semantic con  embedding:动机：对输入信息根据不同的格式进行embedding，并且根据结构建的相关性，计算对应的相似度。
4：任何的embdding都会导致信息的丢失，原始的文本
5：MOE+GATE NET
6:OUTPUT: 0/1

4.2 Input Layer
resume A, B
resume A, B 抽取：结构化：包含xxx，
      chunks，
如何分割产生chunk的motivatin和方法：工作时间段作为分割依据

4.3 embedding
4.3.1 augmentation/finetune
mask: motivation 数据集skew,咋做的

sparse，dense
contrastive loss 微调

4.3.2：motivation。细节
motivation：
输入：
输出：
公式：

4.3.3
4.3.4

\fi

%\begin{CJK}{UTF8}{gbsn}

\begin{figure*}[t]
  \includegraphics[width=\textwidth]{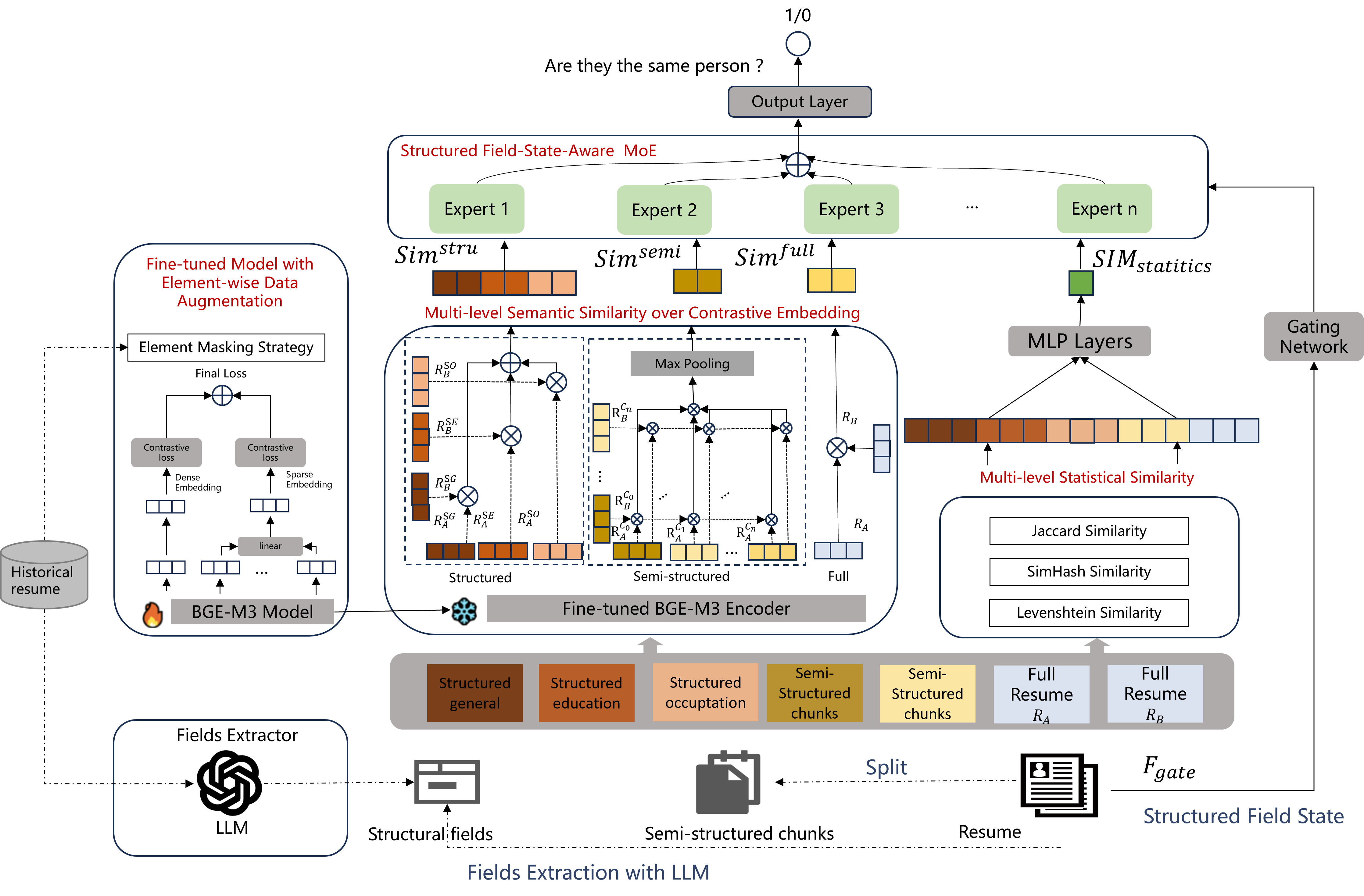}
  \caption{System architecture of \modelname{}.
  %{\color{red} structural fields input,CAn, CBm, RA, RB no max pooling, offline fine-tune linear include all token}
}
\label{fig:system}
\end{figure*}

\subsection{Problem Statement and System Overview} 

{\definition Duplicate Resume Detection Problem} 
Given two resumes $R_A$, $R_B$, the duplicate resume detection problem is to calculate the similarities between $R_A$ and $R_B$, so as to identify whether $R_A$ and $R_B$ belong to the same person.
Specifically, a resume $R_A = \{R_A^{SG}, R_A^{SE}, R_A^{SO},R_A^{C}\}$,
where $R_A^{SG}, R_A^{SE}, R_A^{SO}$ are structured general, education, and occupation information, respectively.
$R_A^{C}$ is a set of semi-structured chunks which are split from the details of involved projects, received awards, and so on.
Structured information refer to the content entered by users within predefined fixed modules of the resume template, while semi-structured information consist of text in open-ended areas of the template that allow flexible input. 

As illustrated in Figure~\ref{fig:system},
to conduct duplicate resume detection, we propose \modelname{}, an MoE-based Hierarchical Semantic Representation Network.
%\modelname{} represents a user's resume as structured, semi-structured, and unstructured information. The structured information includes educational background, personal basic information, and work experience. The semi-structured information involves segmenting the resume into multiple chunks, such as dividing the complete resume based on the user's work experience timeline. The unstructured information is the complete raw resume.
Firstly, to deal with the semantic nature of resumes, \modelname{} fine-tunes \usedllm{}~\cite{bgem3} to achieve more accurate embedding. Specifically, element masking strategies are utilized to augment samples for better fine-tuning.
%
%However, due to the sparsity, imbalance of resume samples, and the lack of negative samples, we propose a strategy of using element masking to generate a series of positive and negative samples for data augmentation, enabling the fine-tuning of the model. Based on the fine-tuned \usedllm{} model, we obtain feature vectors for the three types of resume structure information.
Secondly, to fully utilize the local and global information in resumes, 
multi-level semantic similarity over contrastive embedding is computed.
In detail, the fine-tuned \usedllm{} is used to get the embeddings for structured, semi-structured and full resumes. And the cosine similarity is calculated over these contrastive embeddings. 
Thirdly, since any embedding may lead to information loss, in order to preserve the original information in resumes, multi-level statistical similarity is computed.
Finally, to deal with diverse missing fields, a state-aware MOE is proposed with a gating network to assign weights to each experts according to the input resumes.
%Therefore, we perform three types of similarity calculations on the original information and generate input information for a gating network through a single MLP Layer.
%\textcolor{blue}{To calculate the similarity between resumes, we further train the model by performing contrastive learning on the three types of extracted information, generating multi-dimensional contrastive result vectors. LU:+++ verify 对比学习这部分应该在finetune bge的地方说，也就是上面自然段} However, since any embedding may lead to information loss, we aim to better preserve the original feature information. Therefore, we perform three types of similarity calculations on the original information and generate input information for a gating network through a single MLP Layer.
%After contrastive learning training, we obtain many different result vectors. To integrate all information into the duplicate resume detection decision, we adopt a gating network approach. We train different experts by assigning weights to them through the Gating Network, and the final result is obtained through a single Output Layer, producing an output of 0 or 1. A value of 0 indicates that the two resumes are not from the same person, while a value of 1 indicates that the two resumes are from the same person.

%\subsection{Input Layer} \label{sec:Input Layer}
%\stitle{Structured Information}
%\stitle{Unstructured Information}
%\stitle{Semi-Structured Information}

\subsection{Fine-tuned Model with Elements Data Augmentation}
%In the process of feature generation, different generation methods can capture various information from the data, reduce overfitting, and enhance the robustness of the model. Therefore, after obtaining $R$, $R^S$ and $R^{Semi}$, we use \usedllm{} model to process the data and generate both dense and sparse feature vectors.
%\usedllm{} is known for its multi-functionality, supporting dense retrieval, sparse retrieval, and multi-vector retrieval. This versatility allows it to capture different aspects of the data effectively. Dense vectors capture semantic meaning, while sparse vectors highlight keyword-based information, which is particularly useful for long documents.
%To further improve the model's performance in duplicate resume detection, we fine-tune the model by generating positive and negative samples. This approach helps the model learn better representations and achieve better performance when processing actual resume data.

\modelname{} fine-tunes \usedllm{} for embedding to achieve better resume representation.
%, as \usedllm{} is known for its multi-functionality, supporting dense retrieval, sparse retrieval, and multi-vector retrieval. This versatility allows it to capture different aspects of the data effectively.
To do this, we first generate more positive and negative sample resumes to enrich the data, and then fine-tuning \usedllm{} with contrastive learning.

%\subsubsection{Element Masks Strategy}
\textbf{Elements Masking Strategy}
is proposed to generate more positive and negative sample resumes for fine-tuning.
Specifically, we generate positive sample resumes through hiding part information in the original resume, and the masking strategy follows the distribution of resumes obtained from third-party.
For negative samples, we generate resume pairs of different individuals, and these selected resumes have some similar structured information, and some similar semi-structured detailed experiences.

%\subsubsection{Fine-tuned Embedding Model}
%{\color{red}LU: +++general framework for fine-tuning;mention that we follow the fine-tuning procedure of \usedllm{} }

\textbf{\usedllm{} Fine Tuning}
To capture semantics in resumes, we introduce rich semantic representations, including both dense and sparse representations.
Moreover, in the face of the sparsity of samples and the limited number of positive and negative samples, we employ contrastive learning to fine-tune the model.
%
%{\color{red} QUESTION: 这部分是不是完全按照\usedllm{}写的？这里三个步骤需要展开写吗，还是一句话一个reference带过？ ANSWER:yes,不需要展开说，但是要强调微调了全部的参数，包括linear层参数}
%{\color{red} ++TY: dense和sparse此处略写,直接介绍loss}
%\stitle{Dense Embedding}
%Dense Embedding can map data into a relatively low-dimensional vector space, capable of capturing continuous feature information and complex semantics, with all dimensions of the vector being non-zero. We draw inspiration from the BERT model proposed by Jacob Devlin et al.{\color{red}LU:++reference}, which is based on the Transformer architecture. The process of generating dense feature vectors in the 
%\stitle{Dense \& Sparse Embedding}
%
%\subsubsection{Contrastive Loss}

Specifically, the \textbf{dense embedding} $v$ is used to capture the meaning of individual tokens and their interrelationships within a sentence.
On the basis of the dense embedding, a learned \textbf{sparse embedding} $v'$ is obtained through a linear layer, and is used for more precise matching and semantic-level similarity retrieval.

%\stitle{Contrastive Loss}

%{\color{red}根据同人判定的特点，我们希望fine tune的效果是相似的样本的表征更接近，不相似的样本表征要更远，匹配对比学习。既有dense又有sprase， L final = L dense + L sparse}

Contrastive learning is utilized to fine-tune \usedllm{} by learning the similarities and differences between samples.
And the \textbf{loss for contrastive learning} depends on whether the samples are positive or not. In detail, 
\begin{equation}
\label{eq:contrative}
    \begin{aligned}
     \mathcal{L}_{pos} & = (1-\textit{cs})^2 \\
     \mathcal{L}_{neg} & = reLU (0, (\textit{cs}-0.2)^2)
    \end{aligned}
\end{equation}
where the $\textit{cs}$ represents the cosine similarity over the embedding vectors of two resumes.
The fine-tuning process can enhance the robustness and the performance of \usedllm{}.
%After the resume data, which consists of positive and negative samples, is represented as features, the model uses cosine similarity for contrastive learning to compute a score. The loss value is then calculated based on whether the compared resume is a positive or negative sample:
%Positive Sample Loss Calculation
%Negative Sample Loss Calculation
%The model performs backpropagation based on the calculated loss to fine-tune itself. This process prepares the model for subsequent duplicate resume detection of resumes and enhances its robustness and performance.

\subsection{Multi-Level Semantic Similarity over Contrastive Embedding} 
To capture local and global semantic similarities between resumes, 
we first extract structured fields from resumes with LLM, and split the semi-structured content in resumes into chunks. 
Then, for each resume, like $R_A$, multi-level sparse and dense contrastive embeddings are extracted with fine-tuned \usedllm{}, including sparse and dense embeddings of structured fields $R_A^{SG}$,$R_A^{SE}$,$R_A^{SO}$, embeddings of semi-structured chunks $R_A^{C_i}$, and embeddings of full resumes $R_A$.
After that, similarity networks are constructed to compute the multi-level semantic similarities. And the details of similarity networks are shown below.

%Specifically, we conduct contrastive embedding over structured fileds, semi-structured chunks
%with \usedllm{}
%To fully calculate the similarity between resumes, we fine-tuned the model by generating positive and negative samples, followed by using the model for duplicate resume detection. We conducted contrastive learning training on the three types of resume information extracted from different dimensions and aspects. This approach allows us to capture various aspects of the data and enhance the model's performance in duplicate resume detection.
%\subsubsection{Structured Similarity Network}

As only information in the same structured field is meaningful to be compared, the \textbf{structured similarity network} computes the cosine similarity of the corresponding structured fields between $R_A$ and $R_B$.
Moreover, to capture rich semantic information, both sparse embedding and dense embedding are used in structured similarity network. As a result, the semantic similarity vector of structured fields is $\textit{sim}^{stru}$ as shown in Equation~\ref{eq:structured}.
%
%After obtaining the structured information of two resumes, the model calculates the cosine similarity for each of its three components: basic information, educational information, and work experience information. Each structured component has corresponding dense and sparse feature vectors. Therefore, after the similarity calculation, six floating-point results are obtained. The model concatenates these into a 6-dimensional vector and outputs it.
%The formula for cosine similarity calculation of structured information is as follows:
\begin{equation}
\label{eq:structured}
    \begin{aligned}
        \textit{sim}^{stru} & = concat (\mathcal{CS}_{SG}^D, \mathcal{CS}_{SG}^S,\mathcal{CS}_{SE}^D, \\
        & \quad \quad \mathcal{CS}_{SE}^S, \mathcal{CS}_{SO}^D, \mathcal{CS}_{SO}^S) \\
        \mathcal{CS}_{SG} & = \frac{ {R}_{A}^{SG} \cdot {R}_{B}^{SG}}{\|\mathbf{{R}_{A}^{SG}}\|\|\mathbf{{R}_{B}^{SG}}\|} 
    \end{aligned}
\end{equation}
where $\mathcal{CS}_{SG}^D, \mathcal{CS}_{SG}^S$ represents the cosine similarity of the structured general information over dense and sparse contrastive embeddings, respectively.
%, ${R}_{A}^{S}$represents the structured information of the first resume, and ${R}_{B}^{S}$ represents the structured information of the second resume.

%\subsubsection{Semi-structured and Unstructured Similarity Network}
Duplicate resumes should have some similar detailed experiences, like project experiences and award experiences.
As the detailed experiences are semi-structured content and are always too long, we split the content into chunks according to specified rules (like fixed length or time intervals).
Because incomplete resumes may lack details of some experiences, the highest similarity between semi-structured chunks is most likely to reflect the real similarity between two resumes. 
Thus, the \textbf{semi-structured similarity network} computes the semantic similarity between each pair of chunks in $R_A$ and $R_B$, and the maximum value is identified through max pooling, as shown in Equation~\ref{eq:semi}.
%
%The semi-structured information consists of chunks extracted from the two resumes, with each resume having a variable number of chunks. The model sequentially selects chunks from the two resumes for cosine similarity calculation. Suppose the first resume has n chunks and the second resume has m chunks; the model will perform n × m similarity calculations and output the maximum value based on the results.
%The selection of the maximum value for cosine similarity calculation of semi-structured information chunks is as follows:
\begin{equation}
\label{eq:semi}
    \begin{aligned}
        \textit{sim}^{semi} &= concat(\mathcal{CS}_{semi}^D, \mathcal{CS}_{semi}^S) \\
        \mathcal{CS}_{semi} & = \max_{\substack{i=1,\ldots,m \\ j=1,\ldots,n}}(\frac{R_A^{C_i} \cdot R_B^{C_j}}{\|\mathbf{{R}_{A}^{C_i}}\|\|\mathbf{{R}_{B}^{C_j}}\|})
    \end{aligned}
\end{equation}
where $\mathcal{CS}_{semi}^D$ represents the maximum cosine similarity among dense-embedded semi-structured chunk pairs in $R_A$ and $R_B$, while $\mathcal{CS}_{semi}^S$ represents the similarity over sparse embeddings.
%calculation result of the semi-structured information of the two resumes, ${C}_{i}^{A}$ represents the i chuck block in the semi-structured information of the first resume, ${C}_{j}^{B}$ represents the j chuck block in the semi-structured information of the first resume, and max() function represents the maximum value of the selected output.

%Since each chunk of semi-structured information has corresponding dense and sparse feature vectors, the similarity calculation yields two floating-point results. The model concatenates these into a 2-dimensional vector and outputs it.

Besides the local similarity computed in $\textit{sim}^{stru}$ and $\textit{sim}^{semi}$, the global similarity between resumes is also important.
Thus, the \textbf{un-structured similarity network} calculate the cosine similarity between full $R_A$ and $R_B$, and the final semantic similarity $\textbf{SIM}_{semantic}$ is obtained using Equation~\ref{eq:full}.

\begin{equation}
\label{eq:full}
    \begin{aligned}
        \textit{sim}^{full} &= concat(\mathcal{CS}_{full}^D,\mathcal{CS}_{full}^S )\\
        \mathcal{CS}_{full}& = \frac{ {R}_{A} \cdot {R}_{B}}{\|\mathbf{{R}_{A}}\|\|\mathbf{{R}_{B}}\|} \\
        \textbf{SIM}_{semantic} &=  concat(\textit{sim}^{stru},\textit{sim}^{semi},\textit{sim}^{full})
    \end{aligned}
\end{equation}

%For the unstructured information, which is the original resume, the model generates dense and sparse feature vectors through embedding. Cosine similarity is then calculated for each of these vectors, resulting in two floating-point values. The model concatenates these into a 2-dimensional vector and outputs it.
%The formula for cosine similarity calculation of unstructured information is as follows:

\subsection{Multi-Level Statistical Similarity} 
%In Section ~\ref{sec:Input Layer}, we introduced various methods for extracting information from resumes and the resulting data. However, 
Although fined-tuned \usedllm{} can provide better embeddings to capture rich semantic information in resumes, 
any embedding may lead to information loss, which in turn affects similarity calculation.
%feature generation and the degree of resume similarity—something we aim to avoid. 
Therefore, we also compute the statistical similarities $\textbf{SIM}_{statistic}$ between $R_A$ and $R_B$ using the complete original resumes.
As shown in Figure~\ref{fig:system}, similar to $\textbf{SIM}_{semantic}$, multi-level similarities are calculated over structured pairs, semi-structured pairs, and full pairs of $R_A$ and $R_B$.
Specifically, 
the Jaccard similarity, SimHash similarity and Levenshtein similarity are utilized as the measures to get a similarity vector $\textit{sim}^{text}\in \mathbb{R}^{15}$.
In order to more comprehensively represent the statistical similarity, an MLP layer is used to calculate as:
\begin{equation}
\label{eq:statistics}
    \textbf{SIM}_{statistic} = \texttt{MLP}(\textit{sim}^{text})
\end{equation}

%$$.
%for a series of similarity comparisons and generating a one-dimensional vector to more comprehensively represent the similarity between two resumes.
%This approach enhances the model's robustness and ability to handle complex information. We hope to generate feature statistics for the original resumes through the following three different similarity calculation methods.

\subsection{Structured Field-State-Aware Mixture-of-Experts}
After obtaining $\textbf{SIM}_{semantic}$ and $\textbf{SIM}_{statistic}$, \modelname{} conduct duplicate detection between $R_A$ and $R_B$.
%不同pair的similarity对最终duplicate detection的影响受原始resume中信息缺失情况而影响。
%为了更好地处理多样的信息缺失情况，我们使用MOE处理semantic similarity 和 statistical similarity。
%
%具体来说，根据公式，gating network使用简历中结构化信息的状态作为输入，如果简历中缺失教育信息，则变量a为1， 否则为0.例如，如果简历a中仅有结构化基本信息与结构化教育信息，简历b中仅有结构化教育信息与结构化工作信息，那么对应的门控输入就是【】。
The impact of similarity between different pairs on final duplicate detection is influenced by the extent of information missing in the original resume.
To better handle diverse missing information, we employ Mixture of Experts (MoE) to process semantic similarity and statistical similarity.

Specifically, according to Equation~\ref{eq:gate}, the \textbf{gating network} takes the status of structured information in the resume as input. If the resume $R_A$ lacks educational information, the variable $R_A^{SE}.isNULL$ is set to 1; otherwise, it is 0. 
%For example, if Resume A contains only basic structured information and structured educational information, while Resume B contains only structured educational information and structured work experience information, the corresponding gating input would be [1, 0, 1, 0].
%
\begin{equation}
\label{eq:gate}
    \begin{aligned}
        \mathcal{F}_{gate} & = linear (R_A^{SG}.isNULL, R_B^{SG}.isNULL\\
        & \quad \quad R_A^{SE}.isNULL, R_B^{SE}.isNULL\\
        & \quad \quad R_A^{SO}.isNULL, R_B^{SO}.isNULL) \\
        \mathcal{G} & = softmax (\mathcal{F}_{gate})
    \end{aligned}
\end{equation}

%The MOE (Mixture of Experts) model is characterized by its flexibility, strong adaptability, and inter-pretability. The weights assigned by the gating network to each expert make the decision-making process of the model more interpretable. After the aforementioned feature processing steps, we obtain an 11-dimensional vector, which we input into the MOE model for processing. The model ultimately generates a 0/1 output, where 0 indicates that the two resumes refers to two different individuals, and 1 indicates that the two resumes refers to the same person.

%\subsubsection{Gating Network for generating weight}
%The absence of basic information, educational information, and work experience information in the structured resume data can significantly impact the model's duplicate resume detection. To address this, we implement a Gating Network to record the missing status of these three types of information. Based on the missing conditions, the Gating Network adjusts the strategy for selecting experts and influences the weights assigned to each expert in the duplicate resume detection.

%{\color{red} WRONG: gate output is probability}

%\begin{equation}
%    \begin{aligned}
 %     \mathcal{P}_{gate} & = \text{softmax} (I_{gate})\\
 %    I_{gate} & = GATE(R_A^S, R_B^S)
%     \end{aligned}
% \end{equation}
% where, $GATE (R_A^S, R_B^S)$ checks whether the structural information $R_A^{SG}$,$R_B^{SG}$,$R_A^{SE}$,$R_B^{SE}$,$R_A^{SO}$ and $R_B^{SO}$ is null or not.

% \subsubsection{Moe model training}
In the \textbf{state-aware MOE}, each expert is an MLP network which independently trains and processes the input feature vector to obtain a corresponding score $E_i$, and the total score is calculated as the weighted sum of $E_i$ according to Equation~\ref{eq:moe}.
%each expert consists of a feed-forward neural network and a trainable gating network. Each expert independently trains and processes the input feature vector to obtain a corresponding score. Subsequently, the final score is generated based on the weights assigned to each expert by the Gating Network.
%
%
\begin{equation}
\label{eq:moe}
    \begin{aligned}
    M_i &= \texttt{MLP} (concate (\textbf{SIM}_{semantic}, \textbf{SIM}_{statistic}))\\
     & \textit{Score}  =\sum_{i=0}^{n} \mathcal{G}_i \cdot M_i 
    \end{aligned}
\end{equation}

%{\color{red} QUESTION: equations of MI, OutputLayer is ??}

\subsection{Output Layer}
The obtained \textit{Score} in MOE is fed into the Output Layer, which ultimately outputs the duplicate resume detection result.
\begin{equation}
    \textit{isDuplicate} = softmax(\textit{Score}) 
\end{equation}

%\end{CJK}

\section{Experiment}\label{sec:exp}
%\begin{CJK}{UTF8}{gbsn}
\subsection{Experiment Settings}

\stitle{Datasets}
%The test dataset used in our study is derived from real application scenarios of \modelname{} and has been manually verified. This dataset consists of 139 samples, including 102 positive samples and 37 negative samples. Positive samples represent pairs of resumes belonging to the same individual, where differences between the two resumes may include, but are not limited to, missing or modified information such as names, educational backgrounds, companies, and job descriptions. Negative samples represent pairs of resumes not belonging to the same individual. To more accurately evaluate the model's effectiveness, we did not simply select two different resumes as negative samples but instead adopted hard negatives. These samples are primarily sourced from online cases that underwent secondary review, including cases where the basic system initially classified them as belonging to the same individual but with low confidence scores, as well as user-reported abnormal cases.
%
We conduct the experiments on a real-world dataset collected by a large company in China.
The dataset is collected for more than 6 months.
There are 183205 resume pairs in the dataset, in which 160127 are positive samples and 23375 are negative samples, where positive samples refer to resumes that are textually dissimilar but belong to the same individual, while negative samples denote resumes with textual similarities that originate from different individuals.
%{\color{red} citation of metrics, positive, negative}
%Note that all sensitive or private information has been filtered out from the data.

\stitle{Evaluation Metrics}
We evaluate the duplicate detection task with three widely used metrics for classification:
Area Under the Receiver Operating Characteristic (AUC), Accuracy, and F1 Score.
%We also include the results for Precision, Recall in Appendix~\ref{sec:appendix}.
Specifically, AUC is a curve drawn with true
positive rate as the ordinate and false positive
rate as the abscissa according to a series of
different two classification methods (boundary value or decision
threshold). AUC denotes the Area Under the receiver operating
characteristic curve over the test set, which is a widely used metric for CTR
prediction. The larger AUC is, the better the duplicate detection prediction
model performs.

\stitle{Baselines}
%We compared \modelname{} with several baseline models, including three traditional similarity calculation methods and multiple text embedding models before and after fine-tuning with contrastive learning.
To provide a comprehensive evaluation of our \modelname{} model, we compare it against both traditional similarity calculation methods (i.e., Jaccard, SimHash and Levenshtein similarity ) and LLM-based methods (i.e., mGTE~\cite{mgte}, ME5~\cite{me5}, BGE-Base~\cite{bgebase} and \usedllm{}~\cite{bgem3}).

%\stitle{traditional similarity calculation methods.}

%\stitle{text embedding model}

\stitle{Implementation Details}
We adopted the \usedllm{}~\cite{bgem3} as the LLM model in this paper.
The training process is divided into two main parts: fine-tuning \modelname{} and subsequent model training.
During fine-tuning, given that it already possesses strong text representation capabilities, we conducted only one epoch of training with a batch size of 8 and a learning rate of $1e^{-5}$.
Moreover, the dense embedding is set to $v\in \mathbb{R}^{250002}$ and the sparse embedding is set to $v'\in \mathbb{R}^{1024}$.
After fine-tuning, we froze \usedllm{} and proceeded with the subsequent training phase. 
During detection model training, we trained for 20 epochs with a batch size of 1024 and a learning rate of $1e^{-3}$. Additionally, in our model design, each gate consists of a linear layer, and each expert is composed of a 3 layer MLP.
%{\color{red} 加一些implementation的细节，特别是系统级别的细节，不仅仅是model的细节}
%\in \mathbb{R}^{1024}
%\in \mathbb{R}^{250002}

\begin{table}[ht]
    \centering
    \small
    \caption{Performance of the proposed and baseline methods for duplicate resume detection, where $*$ indicates the best result among baselines. Improvement refers to the enhancement achieved by \modelname{}. }
    \label{tab:overallexp}
    \renewcommand{\arraystretch}{1.2}
    \begin{tabular}{l|c|c|c}
        \hline
        Models & Accuracy & AUC & F1 \\
        \hline \hline
        Jaccard & 0.5036 &  0.5532 & 0.5036 \\
       
        SimHash & 0.4460 & 0.6025 & 0.4296 \\
       
        Levenshtein & 0.7050 &  0.4265 & 0.8194 \\
        
        ME5-base & 0.7554 &  0.7973 & 0.8211 \\
        
        BGE-Base-zh-v1.5 & 0.7122 &  0.7871 & 0.7727 \\
        
        mGTE-Base & 0.7770 &  0.8048 & 0.8394 \\
        
        \usedllm{}(only dense) & 0.7410 & 0.8317 & 0.8043 \\
        
        \usedllm{}(only sparse) & 0.8417 &  0.8486 & 0.8922 \\
        
        \usedllm{}(dense+sparse) & $0.8489^*$ & $0.8593^*$ & $0.8976^*$  \\
       
       \textbf{Ours} \modelname{} & \textbf{0.8849} &  \textbf{0.9096} & \textbf{0.9223} \\
       \hline
       \textbf{Improvement} & $\textbf{5.13\%}$ & $\textbf{7.19\%}$ & $\textbf{3.37\%}$\\ 
        \hline
    \end{tabular}
\end{table}
\subsection{Evaluation Results}
\stitle{Overall Performance}
As shown in Table~\ref{tab:overallexp}, our proposed \modelname{} outperforms all baseline solution, verifying the effectiveness of \modelname{}.
%Moreover, we find that \usedllm{}(dense+sparse) perform best except \modelname{}.
%The results indicate that using both sparse and dense embeddings can further improve the representations of rich context in resumes.
Among the baseline models, \usedllm{} generally outperform traditional similarity calculation methods, while  \usedllm{}(dense+sparse) with supervised contrastive learning perform better than those trained with unsupervised learning. 
This suggests that models fine-tuned with contrastive learning can effectively compute text similarity, leading to more accurate resume similarity judgments.
Moreover, among all fine-tuned models, \usedllm{} performed exceptionally well, which is the main reason for choosing \usedllm{} for fine-tuning. 

%model and presented the model performance in Table \ref{tab:Overall Performance of Different Models on Dataset}. Overall, our model outperforms the baseline models across all metrics. Among the baseline models, text embedding models generally outperform traditional similarity calculation methods, while text embedding models fine-tuned with supervised contrastive learning perform better than those trained with unsupervised learning. This suggests that models fine-tuned with contrastive learning can effectively compute text similarity, leading to more accurate resume similarity judgments. We noticed that among all the fine-tuned models, bge-m3 performed exceptionally well, which is the main reason for choosing this model for fine-tuning. Additionally, we compared the performance of using dense, sparse, and a combination of both dense and sparse embeddings in the bge-m3 model. The results indicate that using both text embedding methods together can further improve the accuracy of determining whether resumes belong to the same person.

%\subsection{Ablation Studies}

\stitle{Ablation Studies}
%{\color{red} 加一些详细的分析，比如哪个新模块最核心？哪个最具有创新性。}
As depicted in Table~\ref{tab:ablation_exp}, all of the components in our proposed \modelname{} have significant impacts on the performance of duplicate resume detection.
In detail, 
removing $\textit{sim}^{semi}$ (Equation~\ref{eq:semi}), $\textit{sim}^{full}$ (Equation~\ref{eq:full}),  and Statistical similarity (Equation~\ref{eq:statistics}) has a minor effect on the \modelname{} performance, with accuracy decreasing by only about 1\%. 
However, after removing $\textit{sim}^{stru}$ (Equation~\ref{eq:structured}) accuracy drops from $88\%$ to $83\%$, suggesting that basic information, educational background, and work experience play a crucial role in resume duplicate detection. 
Compared to Dense embeddings, Sparse embeddings, through the added linear layer that calculates token weights in the text, play a critical role in resume duplicate detection, particularly in structured comparison. 
Furthermore, the accuracy of the model without fine-tuning dropped by $6\%$, while fine-tuning significantly improved the accuracy, confirming the effectiveness of our contrastive learning-based fine-tuning approach. Finally, after removing the MOE component, accuracy decreased by $3\%$, indicating that MOE plays a key role in determining duplicate resume detection.
\begin{table}[ht]
    \centering
    \small
    \caption{Performance of the Ablation Studies.}
    \label{tab:ablation_exp}
    \renewcommand{\arraystretch}{1.2}
    \begin{tabular}{l|c|c|c}
        \hline
        Model & Accuracy &  AUC & F1 \\
        \hline \hline
        w/o Dense embedding & 0.8345 & 0.9059 & 0.8844 \\
        
        w/o Sparse embedding & 0.7985 &  0.8911 & 0.8494 \\
       
        w/o Fine-tuning & 0.8201 &  0.8893 & 0.8663 \\
       
        w/o $\textit{sim}^{stru}$  & 0.8345 &  0.9024 & 0.8832 \\
       
        w/o $\textit{sim}^{semi}$ & 0.8705 &  0.8937 & 0.9108 \\
        
        w/o $\textit{sim}^{full}$ & 0.8705 &  0.8979 & 0.9126 \\
        
        w/o Statistical similarity & 0.8776 &  0.9091 & 0.9154 \\
        
        w/o MoE & 0.8561 & 0.8945 & 0.9009 \\
        \hline
        \modelname{} & \textbf{0.8849} &  \textbf{0.9096} & \textbf{0.9223} \\
        \hline
    \end{tabular}
\end{table}

\stitle{Effect of Chunk Division Methods}
%
%As shown in Table \ref{tab:Case Study of Different Chunk Division Methods}, we experimented with three different chunk division methods, including fixed-length division (with a length set to 512), division by newline characters, and division by time intervals. Compared to the other two methods, dividing by time intervals achieved better performance in terms of accuracy, AUC, and F1 score. This suggests that dividing by time intervals can better capture the temporal dependencies in the sequence, thereby improving the overall performance of the model. Although the precision is lower, this is mainly attributed to the imbalance in the sample distribution.
%
Table~\ref{tab:chunk_exp} illustrates that constructing semi-structured chunks according to time intervals can achieve better performance, as this division method can better capture the temporal dependencies in the sequence.
\begin{table}[ht]
    \centering
    \small
    \caption{Effect of Chunk Division Methods}
    \label{tab:chunk_exp}
    \renewcommand{\arraystretch}{1.2}
    \begin{tabular}{l|c|c}
        \hline
        Chunk Division Mode & AUC & F1 \\
        
        Fixed length(length=512) &  0.9064 & 0.9162 \\
       
        Divided by line break &  0.9069 & 0.9154 \\
       
        Divided by time interval & \textbf{0.9096} & \textbf{0.9223} \\
        \hline
    \end{tabular}
\end{table}

%\stitle{Effect of Num. of Experts}
%\begin{table}[ht]
%    \centering
%    \small
%    \caption{Effect of the Experts Number}
%    \label{tab:Case Study of Different Numbers of Experts}
%    \begin{tabular}{l|c|c}
%        \hline
%        num of Experts &AUC & F1 \\
%        \hline
%        1 & 0.8945 & 0.9009 \\
%        \hline
%        2 & 0.9001 & 0.9054 \\
%        \hline
%        3 & 0.8945 & 0.9064 \\
%        \hline
%        4 & 0.9077 & 0.917 \\
%        \hline
%        5 & 0.9038 & 0.91 \\
%        \hline
%        6 & 0.9096 & 0.9223 \\
%        \hline
%        7 & 0.9075 & 0.9223 \\
%        \hline
%%        8 & 0.9051 & 0.9179 \\
%        \hline
%        9 & 0.9099 & 0.9223 \\
%        \hline
%        10 & 0.9062 & 0.9171 \\
%        \hline
%    \end{tabular}
%\end{table}

\begin{figure}[t]
\centering
%\vspace{-10pt}
\includegraphics[width=0.95\columnwidth]{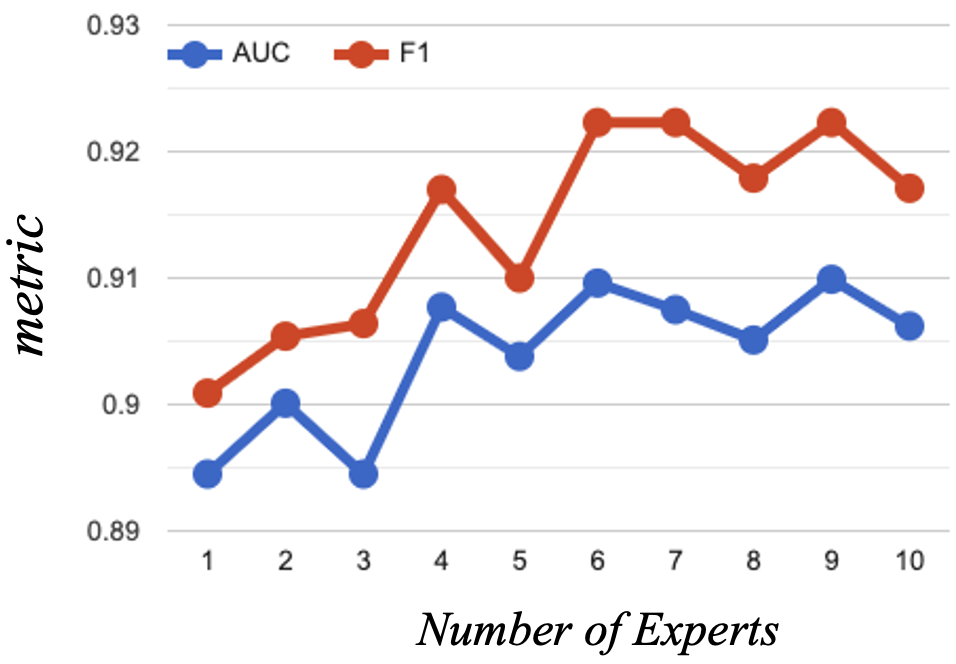}
%\vspace{-5pt}
  \caption{Effect of Number of Experts.  }
\label{fig:exp-numexperts}
\end{figure}

\stitle{Effect of Number of Experts}
Figure~\ref{fig:exp-numexperts} 
%shows that when the number of experts is 6, the model performs well, and thus we adopted this setting in subsequent experiments.
investigates the impact of the number of experts on the judgment results. We evaluate the performance of \modelname{} with varying numbers of experts, from 1 to 10. The experimental results show that when the number of experts is 6, the model performs well over both AUC and F1 metrics. Therefore, we adopted this setting in subsequent experiments.
%

%{\color{red} 描述长一些}

\begin{figure}[!th]
\centering
\includegraphics[width=0.99\columnwidth]{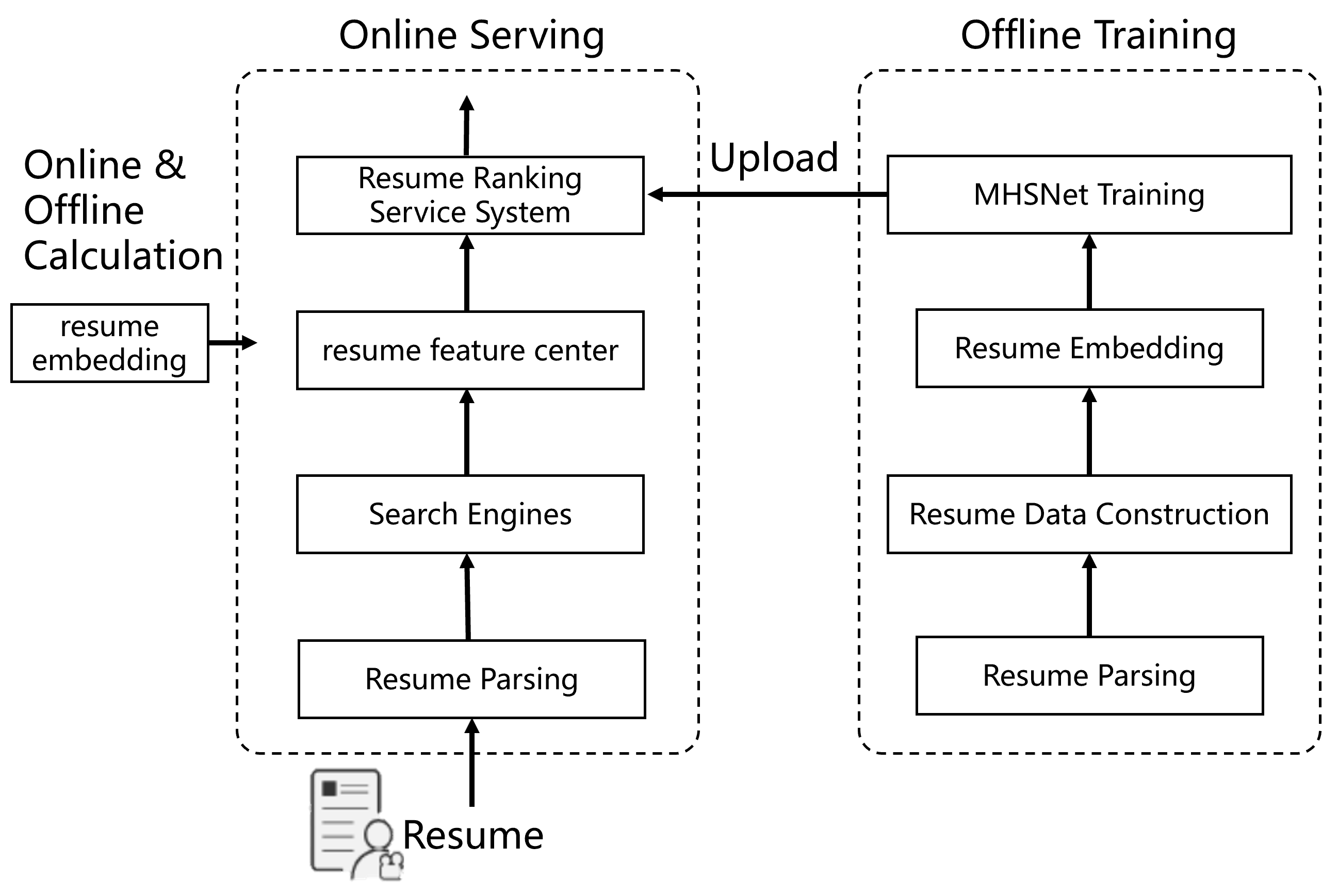}
\caption{The Duplicate Resume Detection Model in Online System. }\label{fig:details}
\end{figure}

\stitle{Online A/B Test}
Figure \ref{fig:details} illustrate the pipeline of online system of duplicate resume detection model (i.e., Step 3 in Figure~\ref{fig:application}).
The online system is deployed on Alibaba's original platform and mainly consists of an online service module and an offline training/inference module. The system leverages a search engine platform to retrieve similar resumes, then uploads the features of these resumes to the online service system, where candidate resumes are re-ranked to determine whether the input resume has duplicates within the internal resume pool.

\begin{figure}[!t]
\centering
%\vspace{-10pt}
\includegraphics[width=0.95\columnwidth]{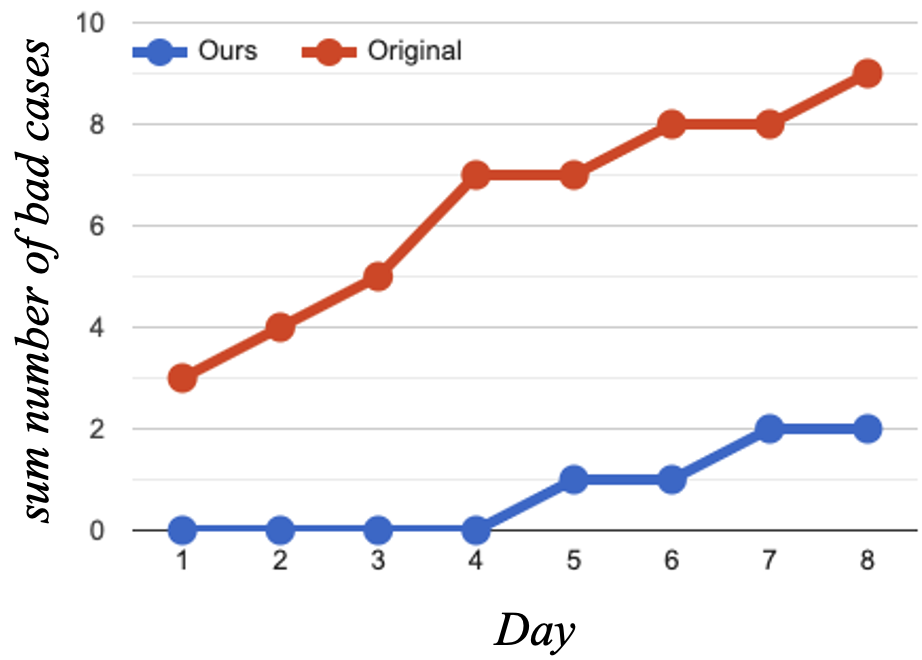}
%\vspace{-10pt}
\caption{Online A/B test.}\label{fig:online_AB_test}
\end{figure}

%\subsection{Online A/B Test}
%\stitle{Online A/B Test}
It is conducted over the real world system used in a large company in China.
We compare the duplicate resume detection performance between the original system and the updated system with \modelname{}.
To ensure fairness, we adjusted the scheduling engine on the online platform so that, during the online A/B tests, approximately half of the daily traffic for duplicate resume detection is allocated to each model.
%Due to the page limit, the pipeline of online system is depicted in Figure~\ref{fig:details} in Appendix~\ref{sec:appendix}.
We monitor the online result for 8 days, and record the accumulated bad cases as shown in Figure~\ref{fig:online_AB_test}.
Here, a bad case refer to a negative sample, which means the system considers two resumes to be from the same person, but in fact, they are not.
According to the result, \modelname{} can improve the online performance.
Moreover, the results also reveal the sparsity and skewness of real-world data, specifically, the scarcity of negative samples.

\begin{figure*}[!t]
\centering
%\vspace{-10pt}
\includegraphics[width=0.8\textwidth]{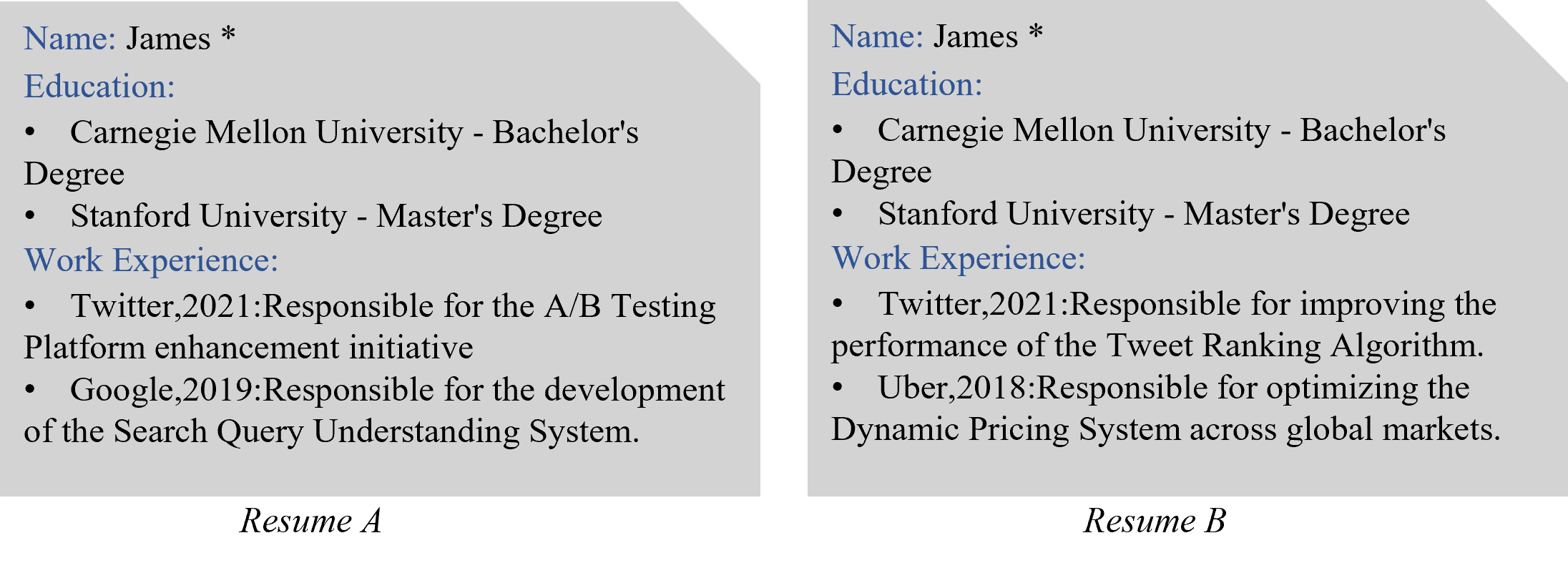}
%\vspace{-10pt}
\caption{Example Resumes.}\label{fig:resumeCase}
\end{figure*}

%\subsection{Case Studies}

%To gain a deeper understanding of the fine-tuning effects of bge-m3, the impact of structured information, and the MOE module on model performance, we designed a series of case studies for evaluation.
%{\color{red} GUAN:+++ 跟着下面的提纲将钉钉文档内容放过来}%

%{\color{red} QUESTION:+++ 这里的case study表达的含义是什么，特别是第一个点}\\
%{\color{blue} ANSWER:+++ 第一个case study对应有无对比学习预训练的case分析，第二个case study对应有无结构化信息的case分析，第三个case study对应有无moe的case分析}

% \begin{table}[ht]
%     \centering
%     \small
%     \caption{Case Study of \usedllm{} Fine Tuning}
%     \label{tab:case-tune}
%     \renewcommand{\arraystretch}{1.2}
%     \begin{tabular}{c|c|c}
%         \hline
%        Text in Structured Field  & with Fine-tuning ? & Similarity \\
%         \hline
%       \makecell{James Michael Smith\\J. M. Smith} & \makecell{YES\\NO} & \makecell{\textbf{0.9732}\\0.7353}\\
%         \hline
%       \makecell{CityU\\Hong Kong City University} & \makecell{YES\\NO} & \makecell{\textbf{0.9488}\\0.8686}\\
%         \hline
%       \makecell{ByteDance \\ TikTok} &  \makecell{YES\\NO} & \makecell{\textbf{0.5}\\0.4907}\\
%         \hline
%     \end{tabular}
% \end{table}

\begin{table}[!t]
    \centering
    \small
    \caption{Case Study of \usedllm{} Fine Tuning}
    \label{tab:case-tune}
    \begin{tabular}{l|c|c|c|c}
        \hline
        \multirow{2}{*}{ Text in Structured Field } & \multicolumn{2}{c|}{no Fine-tuning} & \multicolumn{2}{c}{with Fine-tuning} \\
        \cline{2-5}
        & $\mathcal{CS}_{S}^D$ & $\mathcal{CS}_{S}^S$  & $\mathcal{CS}_{S}^D$ & $\mathcal{CS}_{S}^S$ \\
        \hline
        \makecell{James Michael Smith\\J. M. Smith} & 0.7353 & 0.5365 & 0.9751 & 0.9732 \\
        \hline
        \makecell{CityU\\Hong Kong City University} & 0.8686 & 0.5157 & 0.9526 & 0.9488 \\
        \hline
        \makecell{ByteDance \\ TikTok} & 0.4907 & 0.5005 & 0.6118 & 0.5 \\
        \hline
    \end{tabular}
\end{table}
%

%{\color{red} 有点简短，可以多一些细致的结果分析，突出文章的贡献和创新性。}
\stitle{Case Study of Verifying Fine-tuning}
A case is depicted in Table~\ref{tab:case-tune} where the similarity in each row refers to $\mathcal{CS}_{SG}^D$ (Line 1 in Table~\ref{tab:case-tune}), $\mathcal{CS}_{SE}^D$ (Line 2) and $\mathcal{CS}_{SO}^D$ (Line 3). The result verify that fine-tuning \usedllm{} is necessary. 
The fine-tuned model indicate significant improvements in both Dense and Sparse scores, with the Sparse score nearly doubling from 0.5365 and 0.5157 to 0.9732 and 0.9488, respectively. 
Even for completely dissimilar name text pairs, the fine-tuned model's dense score increased by about $20\%$. These results clearly indicate that contrastive learning fine-tuning significantly enhances the model's ability to capture implicit semantic relationships in resumes, such as name similarity (full names and abbreviation), school and company abbreviation normalization, and other multi-level semantic similarities, thereby improving the overall performance of the model.

%First, we compared three semantically identical text pairs and analyzed the dense and sparse scores of the fine-tuned and non-fine-tuned bge-m3 models. The results are shown in Table \ref{tab:Case Study of Models With and Without Contrastive Learning Fine-Tuning}. The experimental results demonstrate that for similar text pairs containing abbreviations or pinyin, the fine-tuned model showed significant improvements in both dense and sparse scores, with the sparse score nearly doubling, specifically improving from 0.5365 and 0.5157 to 0.9732 and 0.9488, respectively. Even for completely dissimilar name text pairs, the fine-tuned model's dense score increased by about 20\%. These results clearly indicate that contrastive learning fine-tuning significantly enhances the model's ability to capture implicit semantic relationships in resumes, such as name similarity (full names and pinyin), school and company abbreviation normalization, and other multi-level semantic similarities, thereby improving the overall performance of the model.

%
\stitle{Case Study for Verifying MOE}
Considering the resumes $R_A$ and $R_C$ in Figure~\ref{fig:application},
the corresponding $\mathcal{F}_{gate} = [0, 1, 1, 1, 1, 1]$ as the name field in $R_A$ is incomplete. In this case, $\mathcal{G} = [0.0131, 0.8710, 0.8538, $ $0.8487, 0.4553, 0.3084]$, and the final weighted score is 0.3597, which is lower than the score of 0.5325 without using MOE. This indicates that MOE can deal with various missing types well.

%in Table \ref{tab:Resume case} as an example, due to the missing name information in Resume A, the gating input for the MOE is the Boolean vector [0, 1, 1, 1, 1, 1], and the output weights are [0.3058, 0.0406, 0.0430, 0.0311, 0.5341, 0.0455]. The scores of each expert are [0.0131, 0.8710, 0.8538, 0.8487, 0.4553, 0.3084], and the final weighted score is 0.3597, which is lower than the score of 0.5325 without using MOE. This indicates that MOE can dynamically assign weights to different experts based on the missing structured information in the resume. For instance, some experts (such as experts 2, 3, and 4) gave higher scores due to high name similarity, but MOE reduced the probability of these experts through the gating network, effectively lowering the final similarity score.

\stitle{Case Study for $\textit{sim}^{stru}$}
%\textcolor{red}{名字缺失，但相似度得分较高}。
%Next, 
We analyze the effectiveness of the structured information network $\textit{sim}^{stru}$(Equation~\ref{eq:structured}). 
Figure~\ref{fig:resumeCase} presents similar resume cases for two different job applicants, where both Resume A and Resume B use fictitious names. 
Both resumes have the same education experience and worked at the same company in 2021, but in different specific roles.
To calculate $\textit{sim}^{stru}$, we first obtain the General similarity between Resume A and B as 0.95, the Education similarity as 0.9964, and the Occupation similarity as 0.2521. 
With $\textit{sim}^{stru}$, the overall similarity is calculated as 0.3597, which is significantly lower than the similarity of 0.7607 without $\textit{sim}^{stru}$. 
The result indicates that the structured information network can effectively extract the similarity of names and educational background, while reducing the similarity score for subtle differences in work experience, thus more accurately distinguishing between the two resumes.

%\end{CJK}
\section{Related Work}\label{sec:related}
%\subsection{Job Recommendation}

\textbf{Job recommendation} has been widely studied \cite{ramanath2018towards,koren2009matrix}. Text-based methods were introduced, to encode job descriptions and resumes\cite{shen2018,qin2018enhancing,zhu2018person} . Behavior-based methods can capture complex user-job interactions \cite{yang2022modeling,he2020lightgcn}. Hybrid models combined both text and behavior data \cite{le2019towards,jiang2020learning,hou2022leveraging}. Recent trends include adversarial training, behavior graphs \cite{bian2020learning,luo2019resumegan}, and LLM-based recommendations~\cite{wu2024exploring,yingpeng2024}.

\textbf{Semantic Similarity} is widely used in text classification\cite{DBLP:journals/corr/Kim14f} and machine translation\cite{zou2013bilingual}.
Transformer-based models such as BERT\cite{devlin2019bertpretrainingdeepbidirectional} have introduced contextual understanding by combining pre-training and fine-tuning. In addition, advances in text embedding models(e.g., E5 \cite{wang2022text}, BGE \cite{xiao2023c}, and Sentence-T5 (ST5) \cite{ni2021sentencet5scalablesentenceencoders}) have improved semantic similarity measurement.Furthermore, to meet the demands of multilingual scenarios, models such as BGE-M3 \cite{bgem3} have been developed.

\section{Conclusion}\label{sec:conclude}
%We propose \modelname{} to detect duplications over incomplete resumes.
%\modelname{} first fine-tunes \usedllm{} with contrastive learning and element-wise data augmentation.
% Then, local and global similarities between resumes are calculated over different part of resumes.
% After that, MOE is utilized to handle diverse incomplete cases.
% Effectiveness of \modelname{} is verified with comprehensive experiments.
% We provide a thorough introduction to the challenges associated with duplicate resume detection, particularly highlighting the varied parts of resumes (structured, semi-structured, and unstructured).
% To overcome the challenges of duplicate resume detection, we propose \modelname{} to detect duplications over incomplete resumes.
% A key strength of the proposed \modelname{} is its meticulous handling of different resume structures, which ensures \modelname{} adapts effectively to diverse input resume formats. 
% Moreover, the adoption of MoE enables specialized experts to handle different resume fields effectively.
%
We provide a thorough analysis of the challenges in duplicate resume detection, with particular emphasis on the structural heterogeneity of resumes (structured, semi-structured, and unstructured components). To address the challenges of duplicate resume detection, we propose \modelname{}, a novel framework for detecting duplicates in incomplete resumes. The key strength of \modelname{} lies in its hierarchical processing of diverse resume structures, ensuring robust adaptation to heterogeneous input formats. Furthermore, the integration of Mixture of Experts (MOE) enables specialized processing of distinct resume fields.
In detail,
\modelname{} first fine-tunes \usedllm{} with contrastive learning and element-wise data augmentation.
Then, local and global similarities between resumes are calculated over different part of resumes.
After that, MOE is utilized to handle diverse incomplete cases.
Effectiveness of \modelname{} is verified with comprehensive experiments.

\section*{Acknowledgments}

This work is supported in part by the Natural
Science Foundation of Zhejiang University
of Science and Technology (No. 2025QN023), 
the Zhejiang Province Key R\&D Program
(No. 2023C01217), the Fundamental Research Funds for the Provincial Universities of Zhejiang (No. GK249909299001-017), the Natural Science Foundation of Zhejiang Province (No. LQ24F020040),
the Graduate Course Development Project of Zhejiang University of
Science and Technology (No. 2024yjskj03), the Ideological
and Political Education Teaching Research Project of Zhejiang University of Science and Technology (No. 2024-ksj3), and Yangtze River Delta Science and Technology Innovation Community Joint Research Project (No. 2022CSJGG1000/2023ZY1068).

\newpage
\section*{GenAI Usage Disclosure}
The authors hereby disclose that no generative AI technologies were used in the creation or writing of this manuscript.

%\appendix
%\input{7-Appendix}

\bibliographystyle{ACM-Reference-Format}
\balance
\bibliography{custom}


\begin{thebibliography}{25}

%%% the Web address.

\ifx \showCODEN    \undefined \def \showCODEN     #1{\unskip}     \fi
\ifx \showISBNx    \undefined \def \showISBNx     #1{\unskip}     \fi
\ifx \showISBNxiii \undefined \def \showISBNxiii  #1{\unskip}     \fi
\ifx \showISSN     \undefined \def \showISSN      #1{\unskip}     \fi
\ifx \showLCCN     \undefined \def \showLCCN      #1{\unskip}     \fi
\ifx \shownote     \undefined \def \shownote      #1{#1}          \fi
\ifx \showarticletitle \undefined \def \showarticletitle #1{#1}   \fi
\ifx \showURL      \undefined \def \showURL       {\relax}        \fi
% invisible to TeX
\providecommand\bibfield[2]{#2}
\providecommand\bibinfo[2]{#2}
\providecommand\natexlab[1]{#1}
\providecommand\showeprint[2][]{arXiv:#2}

\bibitem[Bian et~al\mbox{.}(2020)]%
        {bian2020learning}
\bibfield{author}{\bibinfo{person}{Shuqing Bian}, \bibinfo{person}{Xu Chen}, \bibinfo{person}{Wayne~Xin Zhao}, \bibinfo{person}{Kun Zhou}, \bibinfo{person}{Yupeng Hou}, \bibinfo{person}{Yang Song}, \bibinfo{person}{Tao Zhang}, {and} \bibinfo{person}{Ji-Rong Wen}.} \bibinfo{year}{2020}\natexlab{}.
\newblock \showarticletitle{Learning to match jobs with resumes from sparse interaction data using multi-view co-teaching network}. In \bibinfo{booktitle}{\emph{Proceedings of the 29th ACM International Conference on Information \& Knowledge Management}}. \bibinfo{pages}{65--74}.
\newblock


\bibitem[Chen et~al\mbox{.}(2024)]%
        {bgem3}
\bibfield{author}{\bibinfo{person}{Jianlyu Chen}, \bibinfo{person}{Shitao Xiao}, \bibinfo{person}{Peitian Zhang}, \bibinfo{person}{Kun Luo}, \bibinfo{person}{Defu Lian}, {and} \bibinfo{person}{Zheng Liu}.} \bibinfo{year}{2024}\natexlab{}.
\newblock \showarticletitle{{M}3-Embedding: Multi-Linguality, Multi-Functionality, Multi-Granularity Text Embeddings Through Self-Knowledge Distillation}. In \bibinfo{booktitle}{\emph{Findings of the Association for Computational Linguistics: ACL 2024}}, \bibfield{editor}{\bibinfo{person}{Lun-Wei Ku}, \bibinfo{person}{Andre Martins}, {and} \bibinfo{person}{Vivek Srikumar}} (Eds.). \bibinfo{publisher}{Association for Computational Linguistics}, \bibinfo{address}{Bangkok, Thailand}, \bibinfo{pages}{2318--2335}.
\newblock
\href{https://doi.org/10.18653/v1/2024.findings-acl.137}{doi:\nolinkurl{10.18653/v1/2024.findings-acl.137}}


\bibitem[Devlin et~al\mbox{.}(2019)]%
        {devlin2019bertpretrainingdeepbidirectional}
\bibfield{author}{\bibinfo{person}{Jacob Devlin}, \bibinfo{person}{Ming-Wei Chang}, \bibinfo{person}{Kenton Lee}, {and} \bibinfo{person}{Kristina Toutanova}.} \bibinfo{year}{2019}\natexlab{}.
\newblock \bibinfo{title}{BERT: Pre-training of Deep Bidirectional Transformers for Language Understanding}.
\newblock
\showeprint[arxiv]{1810.04805}~[cs.CL]
\urldef\tempurl%
\url{https://arxiv.org/abs/1810.04805}
\showURL{%
\tempurl}


\bibitem[Du et~al\mbox{.}(2024)]%
        {yingpeng2024}
\bibfield{author}{\bibinfo{person}{Yingpeng Du}, \bibinfo{person}{Di Luo}, \bibinfo{person}{Rui Yan}, \bibinfo{person}{Xiaopei Wang}, \bibinfo{person}{Hongzhi Liu}, \bibinfo{person}{Hengshu Zhu}, \bibinfo{person}{Yang Song}, {and} \bibinfo{person}{Jie Zhang}.} \bibinfo{year}{2024}\natexlab{}.
\newblock \showarticletitle{Enhancing Job Recommendation through LLM-Based Generative Adversarial Networks}. In \bibinfo{booktitle}{\emph{Thirty-Eighth {AAAI} Conference on Artificial Intelligence, {AAAI} 2024, Thirty-Sixth Conference on Innovative Applications of Artificial Intelligence, {IAAI} 2024, Fourteenth Symposium on Educational Advances in Artificial Intelligence, {EAAI} 2014, February 20-27, 2024, Vancouver, Canada}}, \bibfield{editor}{\bibinfo{person}{Michael~J. Wooldridge}, \bibinfo{person}{Jennifer~G. Dy}, {and} \bibinfo{person}{Sriraam Natarajan}} (Eds.). \bibinfo{publisher}{{AAAI} Press}, \bibinfo{pages}{8363--8371}.
\newblock


\bibitem[He et~al\mbox{.}(2020)]%
        {he2020lightgcn}
\bibfield{author}{\bibinfo{person}{Xiangnan He}, \bibinfo{person}{Kuan Deng}, \bibinfo{person}{Xiang Wang}, \bibinfo{person}{Yan Li}, \bibinfo{person}{Yongdong Zhang}, {and} \bibinfo{person}{Meng Wang}.} \bibinfo{year}{2020}\natexlab{}.
\newblock \showarticletitle{Lightgcn: Simplifying and powering graph convolution network for recommendation}. In \bibinfo{booktitle}{\emph{Proceedings of the 43rd International ACM SIGIR conference on research and development in Information Retrieval}}. \bibinfo{pages}{639--648}.
\newblock


\bibitem[Hou et~al\mbox{.}(2022)]%
        {hou2022leveraging}
\bibfield{author}{\bibinfo{person}{Yupeng Hou}, \bibinfo{person}{Xingyu Pan}, \bibinfo{person}{Wayne~Xin Zhao}, \bibinfo{person}{Shuqing Bian}, \bibinfo{person}{Yang Song}, \bibinfo{person}{Tao Zhang}, {and} \bibinfo{person}{Ji-Rong Wen}.} \bibinfo{year}{2022}\natexlab{}.
\newblock \showarticletitle{Leveraging search history for improving person-job fit}. In \bibinfo{booktitle}{\emph{International Conference on Database Systems for Advanced Applications}}. Springer, \bibinfo{pages}{38--54}.
\newblock


\bibitem[Jiang et~al\mbox{.}(2020)]%
        {jiang2020learning}
\bibfield{author}{\bibinfo{person}{Junshu Jiang}, \bibinfo{person}{Songyun Ye}, \bibinfo{person}{Wei Wang}, \bibinfo{person}{Jingran Xu}, {and} \bibinfo{person}{Xiaosheng Luo}.} \bibinfo{year}{2020}\natexlab{}.
\newblock \showarticletitle{Learning effective representations for person-job fit by feature fusion}. In \bibinfo{booktitle}{\emph{Proceedings of the 29th ACM International Conference on Information \& Knowledge Management}}. \bibinfo{pages}{2549--2556}.
\newblock


\bibitem[Kim(2014)]%
        {DBLP:journals/corr/Kim14f}
\bibfield{author}{\bibinfo{person}{Yoon Kim}.} \bibinfo{year}{2014}\natexlab{}.
\newblock \showarticletitle{Convolutional Neural Networks for Sentence Classification}.
\newblock \bibinfo{journal}{\emph{CoRR}}  \bibinfo{volume}{abs/1408.5882} (\bibinfo{year}{2014}).
\newblock
\showeprint[arXiv]{1408.5882}
\urldef\tempurl%
\url{http://arxiv.org/abs/1408.5882}
\showURL{%
\tempurl}


\bibitem[Koren et~al\mbox{.}(2009)]%
        {koren2009matrix}
\bibfield{author}{\bibinfo{person}{Yehuda Koren}, \bibinfo{person}{Robert Bell}, {and} \bibinfo{person}{Chris Volinsky}.} \bibinfo{year}{2009}\natexlab{}.
\newblock \showarticletitle{Matrix factorization techniques for recommender systems}.
\newblock \bibinfo{journal}{\emph{Computer}} \bibinfo{volume}{42}, \bibinfo{number}{8} (\bibinfo{year}{2009}), \bibinfo{pages}{30--37}.
\newblock


\bibitem[Le et~al\mbox{.}(2019)]%
        {le2019towards}
\bibfield{author}{\bibinfo{person}{Ran Le}, \bibinfo{person}{Wenpeng Hu}, \bibinfo{person}{Yang Song}, \bibinfo{person}{Tao Zhang}, \bibinfo{person}{Dongyan Zhao}, {and} \bibinfo{person}{Rui Yan}.} \bibinfo{year}{2019}\natexlab{}.
\newblock \showarticletitle{Towards effective and interpretable person-job fitting}. In \bibinfo{booktitle}{\emph{Proceedings of the 28th ACM international conference on information and knowledge management}}. \bibinfo{pages}{1883--1892}.
\newblock


\bibitem[Luo et~al\mbox{.}(2019)]%
        {luo2019resumegan}
\bibfield{author}{\bibinfo{person}{Yong Luo}, \bibinfo{person}{Huaizheng Zhang}, \bibinfo{person}{Yonggang Wen}, {and} \bibinfo{person}{Xinwen Zhang}.} \bibinfo{year}{2019}\natexlab{}.
\newblock \showarticletitle{Resumegan: an optimized deep representation learning framework for talent-job fit via adversarial learning}. In \bibinfo{booktitle}{\emph{Proceedings of the 28th ACM international conference on information and knowledge management}}. \bibinfo{pages}{1101--1110}.
\newblock


\bibitem[Ni et~al\mbox{.}(2022)]%
        {ni2021sentencet5scalablesentenceencoders}
\bibfield{author}{\bibinfo{person}{Jianmo Ni}, \bibinfo{person}{Gustavo Hernandez~Abrego}, \bibinfo{person}{Noah Constant}, \bibinfo{person}{Ji Ma}, \bibinfo{person}{Keith Hall}, \bibinfo{person}{Daniel Cer}, {and} \bibinfo{person}{Yinfei Yang}.} \bibinfo{year}{2022}\natexlab{}.
\newblock \showarticletitle{Sentence-T5: Scalable Sentence Encoders from Pre-trained Text-to-Text Models}. In \bibinfo{booktitle}{\emph{Findings of the Association for Computational Linguistics: ACL 2022}}, \bibfield{editor}{\bibinfo{person}{Smaranda Muresan}, \bibinfo{person}{Preslav Nakov}, {and} \bibinfo{person}{Aline Villavicencio}} (Eds.). \bibinfo{publisher}{Association for Computational Linguistics}, \bibinfo{address}{Dublin, Ireland}, \bibinfo{pages}{1864--1874}.
\newblock
\href{https://doi.org/10.18653/v1/2022.findings-acl.146}{doi:\nolinkurl{10.18653/v1/2022.findings-acl.146}}


\bibitem[Qin et~al\mbox{.}(2018)]%
        {qin2018enhancing}
\bibfield{author}{\bibinfo{person}{Chuan Qin}, \bibinfo{person}{Hengshu Zhu}, \bibinfo{person}{Tong Xu}, \bibinfo{person}{Chen Zhu}, \bibinfo{person}{Liang Jiang}, \bibinfo{person}{Enhong Chen}, {and} \bibinfo{person}{Hui Xiong}.} \bibinfo{year}{2018}\natexlab{}.
\newblock \showarticletitle{Enhancing person-job fit for talent recruitment: An ability-aware neural network approach}. In \bibinfo{booktitle}{\emph{The 41st international ACM SIGIR conference on research \& development in information retrieval}}. \bibinfo{pages}{25--34}.
\newblock


\bibitem[Ramanath et~al\mbox{.}(2018)]%
        {ramanath2018towards}
\bibfield{author}{\bibinfo{person}{Rohan Ramanath}, \bibinfo{person}{Hakan Inan}, \bibinfo{person}{Gungor Polatkan}, \bibinfo{person}{Bo Hu}, \bibinfo{person}{Qi Guo}, \bibinfo{person}{Cagri Ozcaglar}, \bibinfo{person}{Xianren Wu}, \bibinfo{person}{Krishnaram Kenthapadi}, {and} \bibinfo{person}{Sahin~Cem Geyik}.} \bibinfo{year}{2018}\natexlab{}.
\newblock \showarticletitle{Towards deep and representation learning for talent search at linkedin}. In \bibinfo{booktitle}{\emph{Proceedings of the 27th ACM international conference on information and knowledge management}}. \bibinfo{pages}{2253--2261}.
\newblock


\bibitem[Shen(2018)]%
        {shen2018}
\bibfield{author}{\bibinfo{person}{et~al. Shen}.} \bibinfo{year}{2018}\natexlab{}.
\newblock \showarticletitle{Joint Representation Learning for Person-Job Fit}. In \bibinfo{booktitle}{\emph{Proceedings of the 27th International Joint Conference on Artificial Intelligence (IJCAI)}}.
\newblock


\bibitem[Wang et~al\mbox{.}(2022)]%
        {wang2022text}
\bibfield{author}{\bibinfo{person}{Liang Wang}, \bibinfo{person}{Nan Yang}, \bibinfo{person}{Xiaolong Huang}, \bibinfo{person}{Binxing Jiao}, \bibinfo{person}{Linjun Yang}, \bibinfo{person}{Daxin Jiang}, \bibinfo{person}{Rangan Majumder}, {and} \bibinfo{person}{Furu Wei}.} \bibinfo{year}{2022}\natexlab{}.
\newblock \showarticletitle{Text embeddings by weakly-supervised contrastive pre-training}.
\newblock \bibinfo{journal}{\emph{arXiv preprint arXiv:2212.03533}} (\bibinfo{year}{2022}).
\newblock


\bibitem[Wang et~al\mbox{.}(2024)]%
        {me5}
\bibfield{author}{\bibinfo{person}{Liang Wang}, \bibinfo{person}{Nan Yang}, \bibinfo{person}{Xiaolong Huang}, \bibinfo{person}{Linjun Yang}, \bibinfo{person}{Rangan Majumder}, {and} \bibinfo{person}{Furu Wei}.} \bibinfo{year}{2024}\natexlab{}.
\newblock \bibinfo{title}{Multilingual E5 Text Embeddings: A Technical Report}.
\newblock
\showeprint[arxiv]{2402.05672}~[cs.CL]
\urldef\tempurl%
\url{https://arxiv.org/abs/2402.05672}
\showURL{%
\tempurl}


\bibitem[Wu et~al\mbox{.}(2024)]%
        {wu2024exploring}
\bibfield{author}{\bibinfo{person}{Likang Wu}, \bibinfo{person}{Zhaopeng Qiu}, \bibinfo{person}{Zhi Zheng}, \bibinfo{person}{Hengshu Zhu}, {and} \bibinfo{person}{Enhong Chen}.} \bibinfo{year}{2024}\natexlab{}.
\newblock \showarticletitle{Exploring large language model for graph data understanding in online job recommendations}. In \bibinfo{booktitle}{\emph{Proceedings of the AAAI Conference on Artificial Intelligence}}, Vol.~\bibinfo{volume}{38}. \bibinfo{pages}{9178--9186}.
\newblock


\bibitem[Xiao et~al\mbox{.}(2023)]%
        {bgebase}
\bibfield{author}{\bibinfo{person}{Shitao Xiao}, \bibinfo{person}{Zheng Liu}, \bibinfo{person}{Peitian Zhang}, {and} \bibinfo{person}{Niklas Muennighoff}.} \bibinfo{year}{2023}\natexlab{}.
\newblock \bibinfo{title}{C-Pack: Packaged Resources To Advance General Chinese Embedding}.
\newblock
\showeprint[arxiv]{2309.07597}~[cs.CL]


\bibitem[Xiao et~al\mbox{.}(2024)]%
        {xiao2023c}
\bibfield{author}{\bibinfo{person}{Shitao Xiao}, \bibinfo{person}{Zheng Liu}, \bibinfo{person}{Peitian Zhang}, \bibinfo{person}{Niklas Muennighoff}, \bibinfo{person}{Defu Lian}, {and} \bibinfo{person}{Jian-Yun Nie}.} \bibinfo{year}{2024}\natexlab{}.
\newblock \showarticletitle{C-Pack: Packed Resources For General Chinese Embeddings}. In \bibinfo{booktitle}{\emph{Proceedings of the 47th International ACM SIGIR Conference on Research and Development in Information Retrieval}} (Washington DC, USA) \emph{(\bibinfo{series}{SIGIR '24})}. \bibinfo{publisher}{Association for Computing Machinery}, \bibinfo{address}{New York, NY, USA}, \bibinfo{pages}{641–649}.
\newblock
\showISBNx{9798400704314}
\href{https://doi.org/10.1145/3626772.3657878}{doi:\nolinkurl{10.1145/3626772.3657878}}


\bibitem[Yang et~al\mbox{.}(2022)]%
        {yang2022modeling}
\bibfield{author}{\bibinfo{person}{Chen Yang}, \bibinfo{person}{Yupeng Hou}, \bibinfo{person}{Yang Song}, \bibinfo{person}{Tao Zhang}, \bibinfo{person}{Ji-Rong Wen}, {and} \bibinfo{person}{Wayne~Xin Zhao}.} \bibinfo{year}{2022}\natexlab{}.
\newblock \showarticletitle{Modeling two-way selection preference for person-job fit}. In \bibinfo{booktitle}{\emph{Proceedings of the 16th ACM Conference on Recommender Systems}}. \bibinfo{pages}{102--112}.
\newblock


\bibitem[Zhang et~al\mbox{.}(2024)]%
        {mgte}
\bibfield{author}{\bibinfo{person}{Xin Zhang}, \bibinfo{person}{Yanzhao Zhang}, \bibinfo{person}{Dingkun Long}, \bibinfo{person}{Wen Xie}, \bibinfo{person}{Ziqi Dai}, \bibinfo{person}{Jialong Tang}, \bibinfo{person}{Huan Lin}, \bibinfo{person}{Baosong Yang}, \bibinfo{person}{Pengjun Xie}, \bibinfo{person}{Fei Huang}, \bibinfo{person}{Meishan Zhang}, \bibinfo{person}{Wenjie Li}, {and} \bibinfo{person}{Min Zhang}.} \bibinfo{year}{2024}\natexlab{}.
\newblock \bibinfo{title}{mGTE: Generalized Long-Context Text Representation and Reranking Models for Multilingual Text Retrieval}.
\newblock
\showeprint[arxiv]{2407.19669}~[cs.CL]
\urldef\tempurl%
\url{https://arxiv.org/abs/2407.19669}
\showURL{%
\tempurl}


\bibitem[Zheng et~al\mbox{.}(2023)]%
        {zhipeng2023}
\bibfield{author}{\bibinfo{person}{Zhi Zheng}, \bibinfo{person}{Zhaopeng Qiu}, \bibinfo{person}{Xiao Hu}, \bibinfo{person}{Likang Wu}, \bibinfo{person}{Hengshu Zhu}, {and} \bibinfo{person}{Hui Xiong}.} \bibinfo{year}{2023}\natexlab{}.
\newblock \showarticletitle{Generative Job Recommendations with Large Language Model}.
\newblock \bibinfo{journal}{\emph{CoRR}}  \bibinfo{volume}{abs/2307.02157} (\bibinfo{year}{2023}).
\newblock
\showeprint[arXiv]{2307.02157}
\urldef\tempurl%
\url{https://doi.org/10.48550/arXiv.2307.02157}
\showURL{%
\tempurl}


\bibitem[Zhu et~al\mbox{.}(2018)]%
        {zhu2018person}
\bibfield{author}{\bibinfo{person}{Chen Zhu}, \bibinfo{person}{Hengshu Zhu}, \bibinfo{person}{Hui Xiong}, \bibinfo{person}{Chao Ma}, \bibinfo{person}{Fang Xie}, \bibinfo{person}{Pengliang Ding}, {and} \bibinfo{person}{Pan Li}.} \bibinfo{year}{2018}\natexlab{}.
\newblock \showarticletitle{Person-job fit: Adapting the right talent for the right job with joint representation learning}.
\newblock \bibinfo{journal}{\emph{ACM Transactions on Management Information Systems (TMIS)}} \bibinfo{volume}{9}, \bibinfo{number}{3} (\bibinfo{year}{2018}), \bibinfo{pages}{1--17}.
\newblock


\bibitem[Zou et~al\mbox{.}(2013)]%
        {zou2013bilingual}
\bibfield{author}{\bibinfo{person}{Will~Y Zou}, \bibinfo{person}{Richard Socher}, \bibinfo{person}{Daniel Cer}, {and} \bibinfo{person}{Christopher~D Manning}.} \bibinfo{year}{2013}\natexlab{}.
\newblock \showarticletitle{Bilingual word embeddings for phrase-based machine translation}. In \bibinfo{booktitle}{\emph{Proceedings of the 2013 conference on empirical methods in natural language processing}}. \bibinfo{pages}{1393--1398}.
\newblock


\end{thebibliography}

\end{document}